\documentclass[letterpaper, 10 pt, conference]{ieeeconf}
\IEEEoverridecommandlockouts 
\usepackage{epsfig}
\usepackage{graphicx}
\usepackage{amsmath}
\usepackage{amssymb}
\usepackage{booktabs}
\usepackage{xcolor}
\usepackage{bbding}
\usepackage{colortbl}
\usepackage{algorithm}
\usepackage{algorithmic}
\usepackage{indentfirst}
\usepackage{multirow}
\usepackage{cite}
\usepackage{url}
\usepackage[colorlinks,linkcolor=blue]{hyperref}

\usepackage[switch]{lineno}

\title{\LARGE \bf
MF-MOS: A Motion-Focused Model for Moving Object Segmentation 

}

\author{ \ \ Jintao~Cheng,
        Kang~Zeng,
        Zhuoxu~Huang,
        Xiaoyu~Tang,
        Jin~Wu,
        Chengxi~Zhang,
        Xieyuanli~Chen,
        Rui~Fan% <-this % stops a space
\thanks{This research was supported by the National Natural Science Foundation of China under Grants 62001173 and 62233013, the Project of Special Funds for the Cultivation of Guangdong College Students’ Scientific and Technological Innovation (``Climbing Program'' Special Funds) under Grants pdjh2022a0131 and pdjh2023b0141, and the Science and Technology Commission of Shanghai Municipal under Grant 22511104500 (\textit{Corresponding author: Xiaoyu Tang}).}
\thanks{Jintao Cheng, Kang Zeng, Xiaoyu Tang are with the School of Electronic and Information Engineering, South China Normal University, Foshan 528225, China. {\tt\small tangxy@scnu.edu.cn}}
\thanks{Zhuoxu Huang is with the Department of Computer Science, Aberystwyth University, Aberystwyth SY23 3DB, U.K. {\tt\small zhh6@aber.ac.uk}}
\thanks{Jin Wu is with the Department of Electronic and Computer Engineering, Hong Kong University of Science and Technology, Hong Kong, China. {\tt\small jin\_wu\_uestc@hotmail.com}}
\thanks{Chengxi Zhang is with the School of Internet of Things Engineering, Jiangnan University, Wuxi, China. {\tt\small dongfangxy@163.com}}
\thanks{Xieyuanli Chen is with the College of Intelligence Science and Technology, National University of Defense Technology, Changsha, China. {\tt\small xieyuanli.chen@nudt.edu.cn}}
\thanks{Rui Fan is with the College of Electronics \& Information Engineering, Shanghai Research Institute for Intelligent Autonomous Systems, the State Key Laboratory of Intelligent Autonomous Systems, and Frontiers Science Center for Intelligent Autonomous Systems, Tongji University, Shanghai 201804, China. {\tt\small rui.fan@ieee.org}}
}

\begin{document}

\maketitle
\thispagestyle{empty}
\pagestyle{empty}

%%%%%%%%%%%%%%%%%%%%%%%%%%%%%%%%%%%%%%%%%%%%%%%%%%%%%%%%%%%%%%%%%%%%%%%%%%%%%%%%
\begin{abstract}
Moving object segmentation (MOS) provides a reliable solution for detecting traffic participants and thus is of great interest in the autonomous driving field. Dynamic capture is always critical in the MOS problem. Previous methods capture motion features from the range images directly. Differently, we argue that the residual maps provide greater potential for motion information, while range images contain rich semantic guidance. Based on this intuition, we propose MF-MOS, a novel motion-focused model with a dual-branch structure for LiDAR moving object segmentation. Novelly, we decouple the spatial-temporal information by capturing the motion from residual maps and generating semantic features from range images, which are used as movable object guidance for the motion branch. Our straightforward yet distinctive solution can make the most use of both range images and residual maps, thus greatly improving the performance of the LiDAR-based MOS task. Remarkably, our MF-MOS achieved a leading IoU of 76.7\% on the MOS leaderboard of the SemanticKITTI dataset upon submission, demonstrating the current state-of-the-art performance. The implementation of our MF-MOS has been released at \href{https://github.com/SCNU-RISLAB/MF-MOS}{https://github.com/SCNU-RISLAB/MF-MOS}.

\end{abstract}

%%%%%%%%%%%%%%%%%%%%%%%%%%%%%%%%%%%%%%%%%%%%%%%%%%%%%%%%%%%%%%%%%%%%%%%%%%%%%%%%
\section{INTRODUCTION}
A key challenge for safe autonomous driving systems is the precise perception of moving objects \textit{e.g.} pedestrians and other vehicles that share the traffic environment \cite{LMnet}. The LiDAR-based MOS task tackles the uncertainty perception in the traffic environment by segmenting the current moving objects such as pedestrians and cyclists while distinguishing a stationary vehicle from one that is moving \cite{MotionSeg3D, Remove,4d}. Therefore, it helps develop such uncertainty perception and is essential in autonomous driving. 
\begin{figure}[t]
\centering
\includegraphics[width=\linewidth]{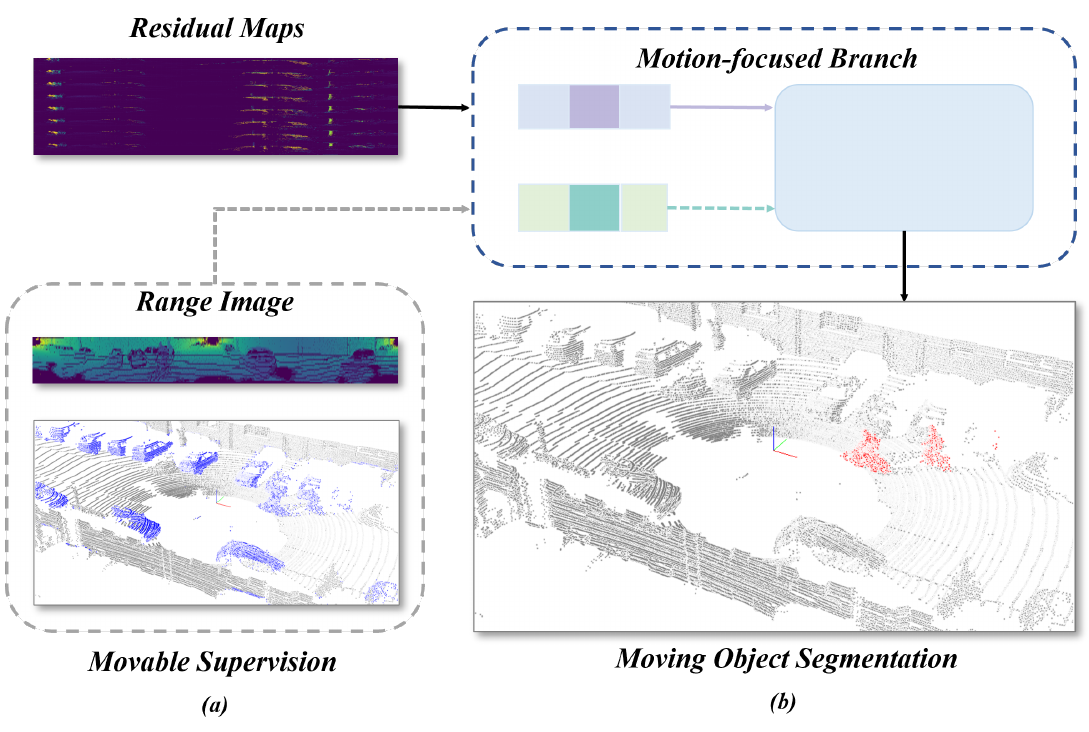}
\caption{Core idea of the proposed motion-focused model. The blue parts in (a) represent the point cloud of movable objects and the red parts in (b) represent the point cloud of moving objects. The moving objects are usually a subset of movable objects. Our MF-MOS emphasizes motion information (via residual maps) and utilizes movable features (via the range image) to provide semantic enhancement. %We observe that moving objects are a subset of movable objects, and the primary task of the proposed method is to identify moving objects. Based on this, our approach emphasizes the extraction of temporal information (via residual maps) and utilizes features of movable objects (via range images) to provide spatial constraints and supervision. In the figure, the blue color represents movable objects, while the red color represents moving objects.
}
\label{motivation}
\vspace{-.5cm}
\end{figure}
This paper follows the mainstream setting of MOS that segments the currently moving objects in a point cloud frame using range projection and residual maps from LiDAR data. Previous methods mainly tackle the dynamic capture with the range view images from point cloud scenes \cite{MotionSeg3D, Auto, LMnet}, while some of them \cite{MotionSeg3D, RVMOS} apply residual maps as auxiliary guidance during dynamic capture.

For instance, MotionSeg3D \cite{MotionSeg3D} utilizes a dual-branch framework based on SalsaNext \cite{SalsaNext}. It simultaneously encodes spatial-temporal information from range images and incorporates a residual branch to enhance motion features. However, these approaches usually prioritize the semantic information of object appearance and detect whether an object can move while relegating the actual motion state of objects to the status of an auxiliary feature. 

Drawing from the most intuitive observations that dynamic capture forms the foundational component in addressing the MOS problem, we put forth MF-MOS, a dual-branch structure for the LiDAR moving object segmentation task. The core idea of the MF-MOS is to focus on the dynamic information from the residual maps as a fundamental component of the network (see Fig. \ref{motivation}). Specifically, we design a primary motion branch to capture the dynamic from residual maps. Additionally, a semantic branch is used to integrate the semantic information of object appearance from range images into the motion branch. We have meticulously designed a kind of pooling layer which is more suitable for the two branches. We name it Strip Average Pooling Layer (SAPL). 

Our method shares a similar envision with RVMOS \cite{RVMOS} that segments movable objects from the range images. However, it primarily emphasizes the motion potential of objects rather than directly addressing their moving status, which is the core aspect of the MOS problem. Different from what, our MF-MOS goes a step further with the direct capture of the dynamic information, thus performing remarkably well in completing the task. More differently, we develop a distribution-based data augmentation to address the influence of different frame sampling of the residual maps to build a robust network. Furthermore, we introduce a 3D Spatial-Guided Information Enhancement Module (SIEM) that provides additional spatial guidance to both the primary motion branch and the semantic branch, thereby alleviating the potential loss of information.

Extensive experiments have demonstrated the superiority of our design. Leveraging the exceptional dynamic perceptual ability of the proposed MF-MOS, we substantially improve the performance of the MOS task on the SemanticKITTI dataset \cite{LMnet} and achieve the top spot on the leaderboard. In summary, our contributions can be summarized as follows:

\begin{itemize}
\item We target the direct capture of the dynamic information in the MOS task and propose a motion-focused network with a dual-branch structure named MF-MOS: (i) a primary motion branch to capture the motion feature from residual maps; (ii) a semantic branch to compute the semantic information of object appearance from range images. 

\item We propose a novel distribution-based data augmentation method that improves the network robustness. We also propose SIEM to refine both the motion branch and alleviate the loss of information.

\item The proposed method attains the highest ranking on the SemanticKITTI-MOS benchmark for both the test and validation datasets. We also tested our method in different benchmarks to validate its robustness and superior performance.
\end{itemize}

\section{Related Work}
\subsection{MOS Based on Occupancy Map and Visibility}
Previous methods address the MOS task by applying occupancy maps or adopting visibility-based methods. They both own the unique advantage of data-free learning. Firstly, inspired by Octomap \cite{hornung2013octomap}, occupancy grids are often utilized in MOS tasks such as moving obstacles removal. These methods compute the motion information by comparing the occupancy maps between continuous frames and locating the dynamic points within the occupancy grid. For instance, J. Schauer et al. \cite{Peopleremover} proposed to remove pedestrians based on the differences in volumetric occupancy between different temporal scans. Likewise, S. Pagad et al. \cite{Robust} targeted to remove dynamic objects on wide urban roads with 3D occupancy maps. Besides, H. Lim et al. \cite{ERASOR} further proposed a pseudo occupancy map based on the height threshold that is robust to motion ambiguity. Secondly, the other type of method adopts the visibility-based theory and applies visual projection data, \textit{e.g.} range images for the MOS task. G. Kim et al. \cite{Remove} proposed a motion points removal and reverting method based on multi-resolution range images. P. Chi et al. \cite{online} proposed a static points construction algorithm via LiDAR and images. However, both the occupancy maps and the visibility-based methods rely on the previously obtained maps and pose information from Simultaneous Localization and Mapping (SLAM) systems, thus being limited in real-time applications.

\subsection{MOS Based on Deep Learning}

Recent approaches tend to apply popular deep learning models and capture spatial-temporal features from data directly. These methods usually adopt different view projections such as range-view projection, voxelization, and bird's eye view projection. For instance, X. Chen et al. \cite{LMnet} presented a novel LiDAR moving segmentation method based on range-view images. Compared to previous approaches, this method prevents static points from being mistakenly removed by capturing canny features. After that, Motionseg3D \cite{MotionSeg3D} was proposed as an improved version. It proposed a dual-branch network with a refined module to optimize the MOS results. RVMOS \cite{RVMOS} also illustrated a multi-branch segmentation framework to fuse semantic and motion information and further improve the MOS performance. %Moreover, it further employs a data augmentation algorithm to make the network solve the slower-moving objects in the test phase. 
Other methods \cite{4d, sparse, wang2023insmos, zhou2023motionbev} adopt different view projections to address the LiDAR-MOS task. B. Mersch et al. \cite{4d} applied the 4D voxelization on LiDAR point clouds for efficiency. Similarly, \cite{zhou2023motionbev} adopted the bird's eye view projection on LiDAR point clouds and proposed a real-time network for the MOS task. We also use the range-view projection in the proposed MF-MOS. Distinct from previous networks, we design a motion-focused network that mainly captures the motion feature from the residual maps and generates semantic features from range images.

\begin{figure*}[htbp]
    \vspace{+.2cm}
    \centering
    \includegraphics[width=0.95\linewidth]{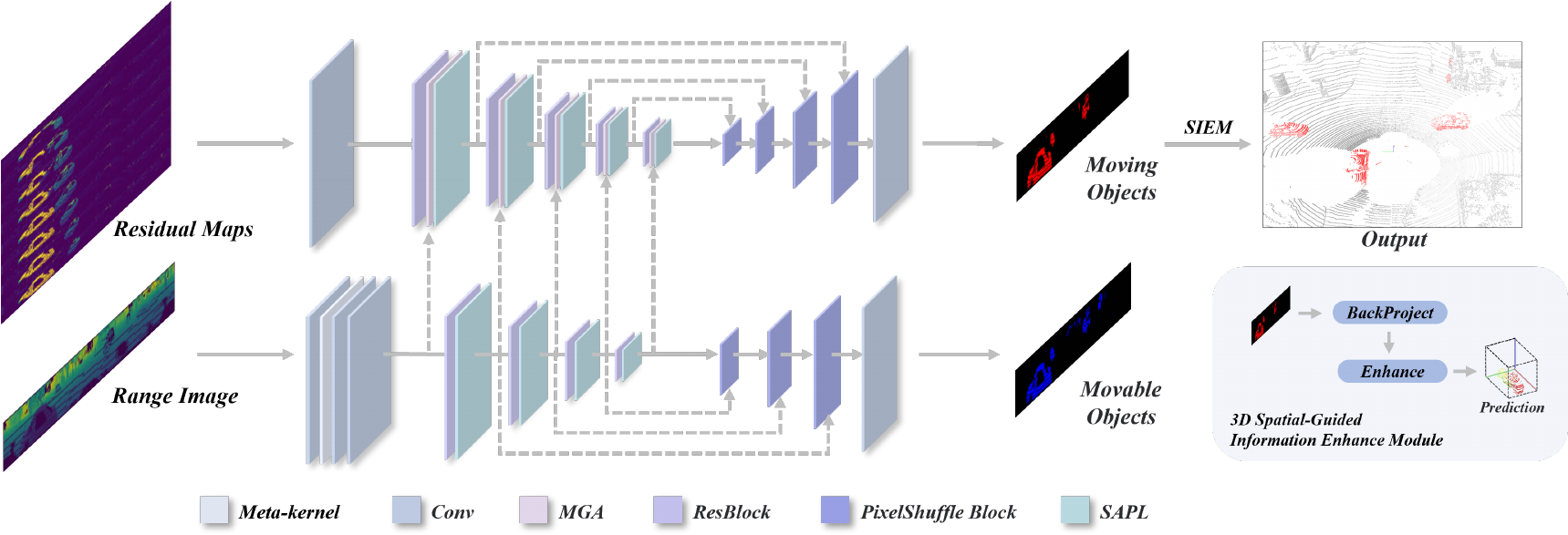}
   %\caption{Dual-Branch, Dual-Head Architecture.  We propose MF-MOS with two branches: an image branch capturing appearance and a residual map branch encoding temporal motion.  These branches merge using a multi-scale motion-guided attention module.  An image decoder with skip connections extracts features, followed by 2D-to-3D projection and further segmentation refinement via a point decoder. We adapt the SalsaNext network by replacing its pooling kernel with the Strip Average Pooling Layer (SAPL), to address dimension misalignment between range and residual maps.}
   \caption{The overall of MF-MOS is a dual-input-dual-output branching structure. The \textbf{semantic branch} (the bottom one) which takes the range image as input is used to predict movable objects in the current frame, and the \textbf{motion branch} (the upper one) takes the residual maps as input to predict the moving objects. The intermediate feature maps obtained from the encoder of the semantic branch are fused into the motion branch through the MGA module. To obtain further refined segmentation results, we use the output of the the motion branch as the input of the SIEM to obtain the final point cloud segmentation results.}
\label{Framework}
\vspace{-.5cm}
\end{figure*}

\section{Methodology}
We present our MF-MOS in detail in the following sections. Firstly, we start with the basic data projection from the LiDAR inputs to the range view and residual inputs. Then, we elaborate on the proposed MF-MOS and the SIEM. Finally, we describe our distribution-based data augmentation for the MOS task in detail.

\subsection{Data Prepossessing}
Range images serve as a lightweight 2D representation of point cloud data. We project the LiDAR point cloud into the range image and residual map following the stander setting of previous work \cite{RVMOS, MotionSeg3D, LMnet}. 
%In 3D space, each point cloud is represented by its spatial coordinates $x$, $y$, and $z$. We then project it to 2D pixel coordinates $u$ and $v$ through sphere mapping, which can be formulated as follows:
%\begin{equation}
%\left(\begin{array}{c}
%u \\
%v
%\end{array}\right)=\left(\begin{array}{c}
%\frac{1}{2}\left[1-\arctan (y, 
%x) \pi^{-1}\right] \quad w \\
%{\left[1-\left(\arcsin \left(z r^{-1}\right)+\mathrm{f}_{\mathrm{up}}\right) \mathrm{f}^{-1}\right] h}
%\end{array}\right),
%\end{equation}
%where  $u$ and $v$ are the transformed pixel coordinates in the 2D image space, $w$ and $h$ are the width and height of the output range images, respectively, $r$ is the range of each point on the 3D space, $f$ is the multi of $f_{\mathrm{up}}$ and $f_{\mathrm{down}}$, which means that the vertical field-of-view of the sensor. 
After getting the range images of different frames of point clouds, we obtain past $k$-frame residual maps $\mathbf{I}_{res}$ by the pixel level variance calculation between frames as follows: 
\begin{equation}
\mathbf{I}_{res}^k(\mathbf{u}, \mathbf{v})=\left|\frac{\mathbf{I}_{R V}^k(\mathbf{u}, \mathbf{v})-\mathbf{I}_{R V}^0(\mathbf{u}, \mathbf{v})}{\mathbf{I}_{R V}^0(\mathbf{u}, \mathbf{v})}\right|,
\end{equation}
where $\mathbf{I}_{R V}$ represents the range image, $u$ and $v$ are the transformed pixel coordinates in the 2D image space.
%\begin{equation}
%d_{k, i}^l=\frac{\left|r_i-r_i^{k \rightarrow l}\right|}{r_i},
%\end{equation}
%Where $d_{k, i}^l$ is the final output residual maps, $r_i$ is the range value of each point from the current frame. All the values are located at image coordinates and ${k \rightarrow l}$ represents the corresponding range value from the transformed scan located at the same image pixel.
\subsection{Network Structure}
\label{Network}

\subsubsection{Dual-Branch Motion-focused Framework with SAPL}
We design a novel dual-branch network that focuses on residual maps. Fig. \ref{Framework} illustrates the overall architecture of the proposed method. Inspired by \cite{LMnet,MotionSeg3D}, SalsaNext has demonstrated powerful performance in the MOS task, hence we employ it as the backbone in motion (via residual maps) and semantic (via the range image) branches.

 The proposed semantic branch utilizes the range image to effectively extract features of movable objects and adds meta kernel \cite{Meta_ker} for feature-level enhancement. In contrast, the feature of the motion branch emphasizes the dynamism of the current features. To enhance the fusion of features from two distinct inputs, we adopt a fusion strategy \cite{MGA} in the motion branch. The encoded feature outputs from the residual maps are combined with those from each layer of the range image, serving as inputs to the subsequent layers of the coding module in the motion branch. This fusion process facilitates the integration of complementary information from both inputs and promotes the effective utilization of features for subsequent processing in the motion branch. The dual-branch framework can be illustrated as:
%Unlike the two-branch input network proposed by \cite{MotionSeg3D, RVMOS} that focuses on range images, we design a two-branch network that focuses on residual maps. Promoted from \cite{MotionSeg3D}, we add meta kernel \cite{Meta_ker} to the semantic branch for feature-level enhancement. To enhance the fusion of features from two distinct inputs, we adopt a fusion strategy in the motion branch. The encoded feature outputs from the residual maps are combined with those from each layer of the range images, serving as inputs to the subsequent layers of the coding module in the motion branch. This fusion process facilitates the integration of complementary information from both inputs and promotes the effective utilization of features for subsequent processing in the motion branch. The dual-branch framework can be illustrated as:
\begin{equation}
\boldsymbol{F}_s=\operatorname{sigmoid}\left(\operatorname{Conv}_{1\times1}\left(\boldsymbol{F}_{\text {semantic }}\right)\right) \otimes \boldsymbol{F}_{\text {motion }}, 
 \label{sigmoid}
\end{equation}
where $\boldsymbol{F}_{\text {motion }}$ represents the feature map of the motion branch in the input fusion module, while $\boldsymbol{F}_{\text {semantic }}$ represents the feature map of the semantic branch in the input fusion module. Followed by 
(\ref{sigmoid}), we can contact two branch features. $\boldsymbol{F}_s$ presents the fusion result after sigmoid processing.
\begin{equation}
\boldsymbol{F}_f=\operatorname{softmax}\left(\operatorname{Conv}_{1\times1}\left(Pool\left(\boldsymbol{F}_s\right)\right)\right) \times C,    
\label{softmax}
\end{equation}
$Pool$ denotes the adaptive average pooling layer, where $C$ represents the number of channels in the feature map. $\boldsymbol{F}_f$ is the output after normalization $softmax$ process.
\begin{equation}    
\boldsymbol{F}_o=\boldsymbol{F}_c+\boldsymbol{F}_{\text {res }},
\end{equation}
where $\boldsymbol{F}_o$ denotes the connection of $\boldsymbol{F}_C$ and $\boldsymbol{F}_{\text {res}}$.
We can note that SalsaNext \cite{SalsaNext} includes a 2$\times$2 pooling layer with a standard square pooling kernel during the feature encoding phase. It is obvious that range and residual maps are generally not aligned in terms of height and width, and the use of a square pooling kernel for feature extraction can easily lead to partial feature loss. Therefore, in the proposed two-branch input network, we modify the pooling layer down-sampled by the SalsaNeXt encoder with the PixelShuffle layer up-sampled by the decoder, which is called the Strip Average Pooling Layer (SAPL), expressed as follows:
\begin{equation}
\boldsymbol{F}_{x_0, y_0}^{\prime}=\frac{1}{h \times w} \sum_{i=0}^w \sum_{j=0}^h \boldsymbol{F}_{w \times x_0+i, h \times y_0+j},
\label{sapl}
\end{equation}
where $\boldsymbol{F}$ represents the input feature map for stripe pooling, while $\boldsymbol{F}^{\prime}$ represents the output feature map after stripe pooling. $h$ and $w$ denote the height and width of the stripe pooling kernel, respectively. $x_o$ and $y_o$ represent the coordinates of the output feature map, corresponding to the $x$ and $y$ dimensions.

In the encoder, we modify the original 2$\times$2 pooling kernel of SalsaNeXt \cite{SalsaNext} by replacing it with a pooling kernel size of 2$\times$4. Additionally, in the PixelShuffle operation, we adjust the ratio of channel-to-image aspect conversion to match the modified rectangular pooling operation.
%On the encoder, the original 2$\times$2 pooling kernel of SalsaNeXt \cite{SalsaNext} will be replaced by a pooling kernel size of 2 $\times$ 4. For the PixelShuffle operation, the ratio of channel-to-image aspect conversion is adjusted to correspond to the modified rectangular pooling operation.

\subsubsection{3D Spatial-Guiding Information Enhancement Module}
%During the process of projecting point clouds onto range images, objects in the point cloud space may overlap in the two-dimensional range view due to the reduction in dimensionality. As a result, in the final stage of range image segmentation, the pixels corresponding to the occluded parts of moving objects will not be accurately segmented by the model.
\begin{figure}[t]
    \centering
    \vspace{+.2cm}
    \includegraphics[width=0.95\linewidth]{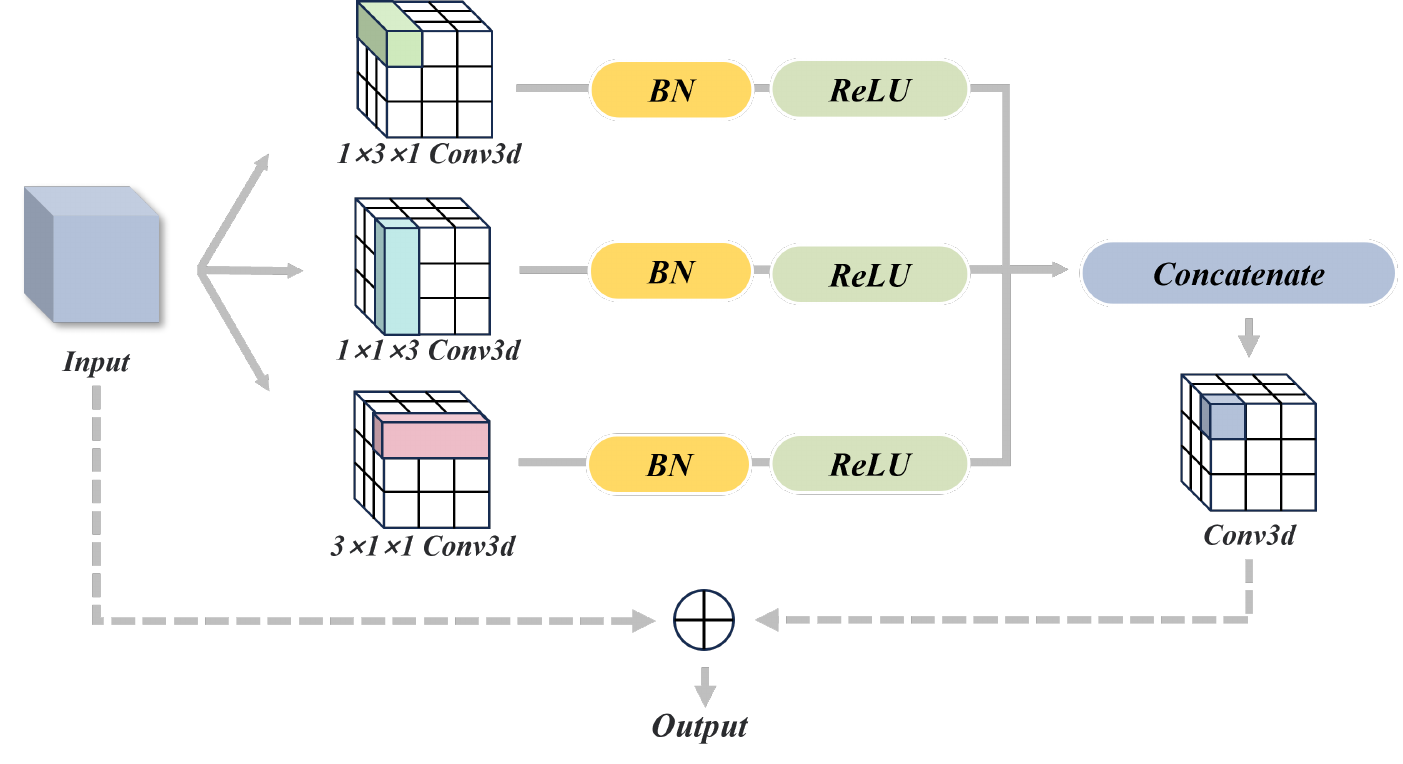}
   \caption{Enhancing 3D Spatial Information with the SGB. The SGB partitions and enriches features across dimensions before fusion, aiming to distill insights from sparse point clouds.}
\label{Block}
\vspace{-.cm}
\end{figure}
To compensate for the information loss caused by the data dimension reduction during the conversion from point clouds to range images, we propose the SIEM to refine the segmentation results of the dual-branch network. SIEM transforms the last layer feature map of the first stage decoder through a back-projection process into the point cloud space, resulting in the initial feature point cloud. After voxelization, the initial point cloud is fed into our proposed 3D Spatial-Guided Block (SGB) for further processing. 
As shown in Fig. \ref{Block}, the input of the SGB module first goes through three different 3D spatial convolution processes to decompose the features into multiple dimensions and enhance the information in each dimension. And then, the outputs from last step are fused by concatenation and 3D convolution in order to capture effective information from the sparse point cloud to a great extent. The final output of the SGB is obtained by skip connection with the original module's input to avoid gradient dispersion.
%\begin{equation}
%\boldsymbol{S}=\left[\operatorname{Conv}^i_{sp}\left(\boldsymbol{F}_I\right)\right], i=1,2,3.
%\end{equation}
%specifically, $\boldsymbol{F}_I$ represents the input of the module, and $\boldsymbol{F}_O$ denotes the output of the module after spatial attention enhancement. In addition, $\operatorname{Conv}_{s}$ represents the spatial convolution operation, while $i$ signifies three different spatial convolution kernels. These symbols are utilized to describe the operations and flow of feature maps within the module.

\begin{figure}[t]
   \centering
   \includegraphics[width=\linewidth]{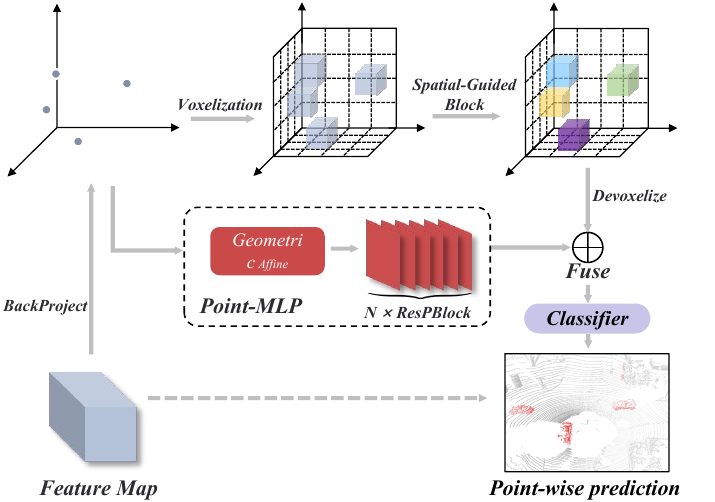}
  \caption{Illustration of the SIEM. The process involves voxelization of the initial feature map, followed by SGB and Devoxelization. The resulting output is fused with the Point-MLP output and classified.
}
\label{2stage}
\vspace{-.5cm}
\end{figure}

Finally, we apply a de-voxelization operation to the processed point cloud and fuse it with the point cloud data after MLP feature extraction. Afterward, a classifier is employed to output the refined per-point segmentation results. Overall, The SIEM can be shown in Fig. \ref{2stage}.

\begin{figure}[t]
    \centering
    \includegraphics[width=\linewidth]{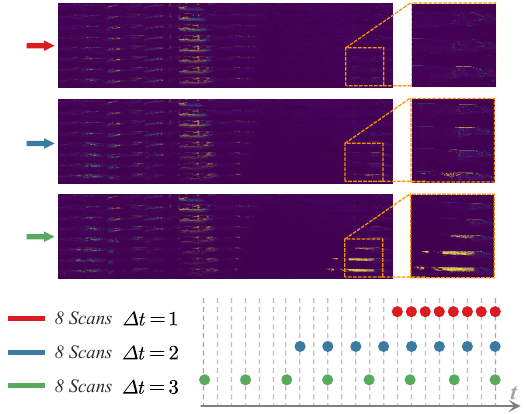}
   \caption{$K$-frames residual maps using different frame stride $\Delta{t}$. The red-boxed region shows residual feature responses correspondence to the different moving speeds of objects. A larger $\Delta{t}$ corresponds to slower-moving objects. Here we show results from eight-frame residual maps.}
\label{aug_compare}
\vspace{-.5cm}
\end{figure}

\subsection{Data Augmentation}
\label{sec:dataaug}
We propose a streamlined yet effective distribution-based data augmentation to improve our MF-MOS. We maintain the idea of motion focus and enhance the learning process in the temporal domain. As shown in Fig. \ref{aug_compare}, different residual maps with different frame strides usually represent different ranges of temporal information. To better enhance the motion features, we propose to generate the residual maps using multiple frame strides instead of a fixed stride. Given a frame stride $\Delta{t} \in [1, 2, 3]$, the correspondence residual maps is represented as $I_{\Delta{t}}$. To prevent data redundancy during training, we augment the data based on the given distributions of the frame strides. In other words, instead of feeding all $I_{\Delta{t}}$ into the network, we choose one $\Delta{t}$ based on a designed distribution probability in every training iteration. We evaluate different distributions and different ranges of the $\Delta{t}$ and report ablation results in Sec. \ref{augmentationAblation}.

\subsection{Loss Function}
During the training process, the total loss function of the proposed algorithm includes the motion-branch losses and the range-branch losses. The sum of the total loss $\mathcal{L}_{\text {Total}}$ can be followed as:
\begin{equation}
\mathcal{L}_{\text {Total}}=\mathcal{L}_{\text {Semantic}}+\mathcal{L}_{\text {Motion}},
\end{equation}
where $\mathcal{L}_{\text {Semantic}}$ represents the losses of the semantic branch with the range image, and $\mathcal{L}_{\text {Motion}}$ is the losses of motion branch with residual maps.

Both of semantic and motion branch used the weighted cross-entropy $\mathcal{L}_{\text {wce}}$ and Lov'asz-Softmax losses $\mathcal{L}_{\text {ls}}$. 
The loss function for each individual branch is as follows:
\begin{equation}
\mathcal{L}=\mathcal{L}_{\text {wce}}+\mathcal{L}_{\text {ls}},
\end{equation}  

%The cross-entropy loss and Lov'asz-Softmax loss are used to supervise both the motion and the semantic branch. In addition, the final loss of the training stage is the sum of the both loss of the motion branch and the semantic branch.

\section{Experiments}
We conduct extensive experiments to comprehensively evaluate our MF-MOS. %We mainly test our method on the most popular and authoritative dataset SemanticKITTI-MOS \cite{LMnet} for the moving object segmentation task. 
In the following sections, we first illustrate the experimental setup for the MOS task, then report the basic validation/test results on the two widely used MOS datasets SemanticKITTI-MOS \cite{LMnet} and Apollo \cite{apollo} to demonstrate the generalization ability of our approach. Following these results, we design our ablation studies fastidiously on the SemanticKITTI-MOS dataset to evaluate the rationality of our MF-MOS. % Please also find the description of the aforementioned datasets and the full training details for reproduction in the \textbf{Appendix}.

\begin{table}[t]
\vspace{+.3cm}
\caption{Comparisons result on SemanticKITTI-MOS dataset.}
\centering
\label{valid}
{\begin{tabular}{lccc}
    \toprule  
    Methods & Publication & Validation (\%) & Test (\%)\\
    \midrule
    SpSequenceNet \cite{spsequencenet} & CVPR 2020 & - & 43.2 \\
    KPConv \cite{kpconv}& ICCV 2019 & - & 60.9\\
    Cylinder3D \cite{Clinder3D} & CVPR 2021  & 66.3 & 61.2 \\
    LMNet \cite{LMnet} & ICRA 2021  & 63.8 & 60.5\\
    4DMOS \cite{4d} & RAL 2022  & 71.9 & 65.2 \\
    MotionSeg \cite{MotionSeg3D} & IROS 2022 &  71.4  & 70.2\\
    RVMOS \cite{RVMOS} & RAL 2022 & 71.2 & 74.7\\
    InsMOS \cite{wang2023insmos} & IROS 2023 & 73.2 & 75.6 \\
    \midrule
    MF-MOS(Ours)  & - & \textbf{76.1} & \textbf{76.7} \\
\bottomrule
\end{tabular}}
\end{table}

\begin{table}[t]
\caption{Comparisons result on Apollo dataset.}
\label{tab:apollo}
\centering
\begin{tabular}{lc}
\toprule  
Methods & IoU (\%) \\
\midrule
MotionSeg3D (\textit{cross-val}) \cite{MotionSeg3D} & 7.5 \\
LMNet (\textit{cross-val}) \cite{LMnet} & 16.9 \\
% LMNet (fully train) \cite{Auto}& 45.7 \\  
LMNet (\textit{fine-tune}) \cite{LMnet} & 65.9 \\
\midrule
MF-MOS (\textit{cross-val}) & 49.9 \\
MF-MOS (\textit{fine-tune}) & \textbf{70.7} \\
\bottomrule
\end{tabular}
\vspace{-.5cm}
\end{table}

\subsection{Experiment Setups}

We utilize the SemanticKITTI-MOS dataset \cite{LMnet} as the main training and evaluating benchmark in our experiment. SemanticKITTI-MOS is the most popular and authoritative dataset for the MOS task with richly labeled moving objects. We use the standard data splits for training, validating, and testing following previous works \cite{LMnet, MotionSeg3D, RVMOS}. As illustrated in Sec. \ref{Network}, we use the semantic labels for movable objects during training to provide additional supervision for the primary motion branch. Additionally, we also perform validation experiments on the Apollo dataset \cite{apollo}, accompanied by some quantitative analysis following the standard experiment setting in \cite{Auto}. 

%  To assess the labeling performance, we rely on the commonly applied Jaccard Index or intersection-over-union (mIoU) metric over moving and non-moving parts of the environment. We map all moving-x classes of the original SemanticKITTI semantic segmentation benchmark to a single moving object class.

%\subsubsection{Datasets}
%SemanticKITTI-MOS \cite{LMnet} is a classical dataset for moving object segmentation tasks with standard labels about moving objects. Similarly to other approaches, we partitioned the dataset into three subsets. 
%Furthermore, we augmented the dataset with semantic labels for movable objects and provided supervision for the main branch. 

%In addition to conducting extensive experiments using the aforementioned dataset, we also performed validation experiments on Apollo \cite{apollo}. Similar to the sequences used in \cite{4d}, a subset of experiments conducted quantitative analysis on the Apollo dataset. 

Our code is implemented in PyTorch. The experiments are conducted on 2 NVIDIA Tesla A100 GPUs. We train our MF-MOS for 150 epochs with an initial learning rate of 0.01 and a decay factor of 0.99 in every epoch. The batch size is set to 8 for each GPU. We use the SGD optimizer with a momentum of 0.9 during training. Following the standard evaluation in MOS, we adopt the intersection over union (IoU) \cite{IoU} of the moving objects to quantify our results in all experiments.

\begin{table}[t]
\vspace{+.3cm}
\caption{Ablation Experiments with Proposed Modules.}
\centering
\label{ablation}
{\begin{tabular}{ccccc}
    \toprule
    Methods & \multicolumn{3}{c}{Component} & IoU (\%) \\
    \midrule
    & Dual-Branch & SIEM & Data Aug &  \\
    LMNet \cite{LMnet} & \multicolumn{3}{c}{} & 63.82 \\
    MF-MOS (\textit{i}) & - & - & - & 64.96  \\
    \midrule
    \multirow{3}{*}{MF-MOS (\textit{ii})} & \Checkmark & - & - & 71.44  \\
    & - & \Checkmark & - & 69.59  \\
    & - & - & \Checkmark & 67.34  \\
    \midrule
    \multirow{4}{*}{MF-MOS (\textit{iii})} & \Checkmark & \Checkmark & - & 74.13 \\
    & \Checkmark & - & \Checkmark & 73.12  \\
    & - & \Checkmark & \Checkmark & 70.47 \\ 
    & \Checkmark & \Checkmark & \Checkmark & \textbf{76.12}  \\
\bottomrule
\end{tabular}}
\end{table}

\begin{table}[t]
\centering
\caption{The proposed modules performance (\%) on other methods.}
\begin{tabular}{cccc}
\toprule 
Method & Baseline & w/ Aug & w/ SIEM   \\
\toprule 
% LMnet & 63.82 & 64.97 & 64.28 & \textbf{65.48}\\
LMNet \cite{LMnet} & 63.82 & \textbf{+1.15} & \textbf{+0.46}  \\
% Motionseg3D-v1 & 68.07 & 68.53 & 71.37 & \textbf{72.10}  \\
MotionSeg3D \cite{MotionSeg3D} & 68.07 & \textbf{+0.46} & \textbf{+3.30}  \\
\bottomrule
\end{tabular}
\label{comparation}
\vspace{-.5cm}
\end{table}

%\subsubsection{Training Details and Evaluation Metric}
%Our model was trained using the PyTorch deep learning framework on 2 NVIDIA Tesla A100 GPUs. The total number of training epochs was set to 150, with an initial learning rate of 0.01. The learning rate was decayed by a factor of 0.99. During training, a batch size of 8 was used on each GPU, and parameter optimization was performed using stochastic gradient descent (SGD) with a momentum of 0.9. During the training process, the total loss function includes the semantic-branch and motion-branch. Both semantic and motion branches used the weighted cross-entropy and Lov'asz-Softmax losses.

\subsection{Comparison with SoTA Methods}
%We first report the validation and test results on the SemanticKITTI-MOS \cite{LMnet}. We use the standard data splits following previous works \cite{}

We first report the validation and test results on the SemanticKITTI-MOS \cite{LMnet} dataset in Tab. \ref{valid}. We achieve state-of-the-art performance on both the validation set and the test set. Remarkably, we improve the validation IoU by 2.9\% compared to the last SoTA \cite{wang2023insmos}. For the test set measurement, we upload our moving object segmentation results to the benchmark server and report the IoU from the leaderboard. Results show that our proposal maintains consistent superiority on the test set. Our performance stays on top and suppresses all the other methods with an IoU of 76.7\%.

We also report the validation result on the Apollo dataset \cite{apollo} in Tab. \ref{tab:apollo}. Following the standard setting of previous approaches \cite{wang2023insmos, Auto}, we adopt protocols including transfer learning and end-to-end fine-tuning for validation experiments on Apollo. The \textit{cross-val} setting refers to cross-validation and the \textit{fine-tune} setting refers to end-to-end fine-tuning. Both pre-train weights are obtained from the SemanticKITTI-MOS training. Our MF-MOS shows significant improvements in both settings.

%, our algorithm exhibits significant improvements compared to LMNet, and MotionSeg3D respectively. When contrasting the retraining and fine-tuning outcomes of LMNet, our algorithm consistently exhibits superior segmentation performance.

%\begin{figure}[H]
%    \centering
%    \includegraphics[width=0.5\textwidth]{figs/Kitti_vis.png}
%   \caption{SemanticKITTI.}
%\label{KITTI}
%\end{figure}

\begin{figure*}[htbp]
    \centering
    \vspace{+.3cm}
    \includegraphics[width=\linewidth]{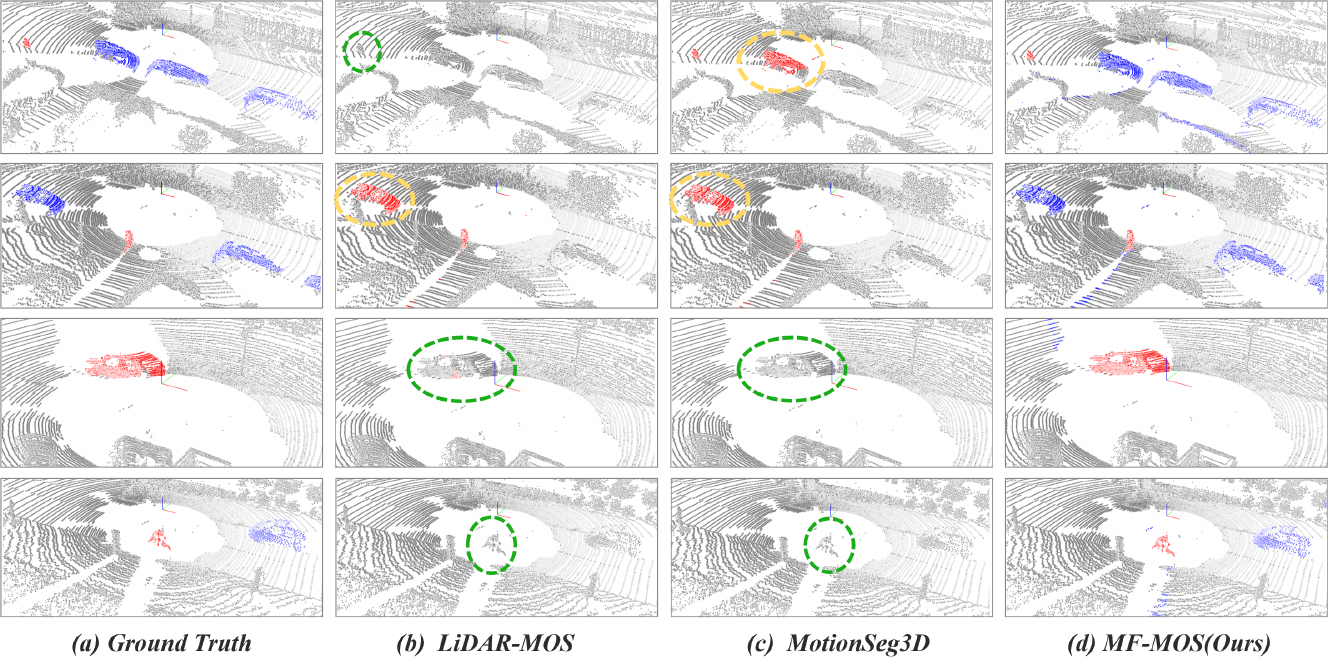}
   \caption{The trinary model visualization comparison exhibits discernible distinctions, where the blue points correspond to movable objects while the red points correspond to moving objects. Those with green circles denote false negatives, and those with yellow circles indicate false positives.}
\label{compare}
\vspace{-.3cm}
\end{figure*}

\subsection{Ablation Studies}
\label{augmentationAblation}
We conduct ablation experiments on the proposed MF-MOS and its different components. The results are shown in Tab. \ref{ablation}. Firstly, without any refinement module, our motion-focused framework proves to be superior. With only the motion branch capturing motion information from the residual maps (setting \textit{i}), MF-MOS exhibits a significant improvement (+1.16\% IoU) compared to LMNet \cite{LMnet}, which uses range images as the main inputs. Additionally, each of the proposed components consistently enhances the performance of our baseline to varying degrees (setting \textit{ii}), with the dual-branch structure alone providing the most significant improvement. To further demonstrate the indispensability of each of our components, we design ablation experiments on different combinations of them in setting \textit{iii}. The last row demonstrates that our full MF-MOS achieved the best performance.

\begin{table}[htbp]
\caption{Ablation Experiments of data augmentation. The sum of distribution probability of $\Delta{t}$ is equal to 1. $\Delta{t}$=max means use the largest frame stride in testing.}
\label{aug}
\centering
\resizebox{\linewidth}{!}
{\begin{tabular}{ccccccccc}
    \toprule
   % \multirow{2}{*}{} & \multirow{2}{*}{Time Intervals} & \multicolumn{5}{c|}{t probability distribution} & \multicolumn{2}{c}{IoU (\%)}\\
   \multicolumn{5}{c}{distribution probability of $\Delta{t}$} & \multicolumn{2}{c}{IoU (\%)} \\
   \midrule
    %\cline{3-9}
    1 & 2 & 3 & 4 & 5 & $\Delta{t}$=1 & $\Delta{t}$=max \\
    \midrule
    % 1 & - & - & - & - & 71.44 & - \\
    0.33 & 0.33 & 0.33 & - & - & 71.41 & 71.91 \\
    0.25 & 0.25 & 0.25 & 0.25 & - & 71.12 & 72.13\\
    0.2 & 0.2 & 0.2 & 0.2 & 0.2 & 68.69 & 70.13 \\
    \midrule
    0.4 & 0.3 & 0.3 & - & - & 70.28 & 71.02 \\
    0.5 & 0.25 & 0.25 & - & - & 71.62 & \textbf{73.12} \\
    0.6 & 0.2 & 0.2 & - & - & 70.52 & 70.51 \\
\bottomrule
\end{tabular}}
\vspace{-.3cm}
\end{table}

To assess the effectiveness of the proposed SIEM and the distribution-based data augmentation at a deeper level, we apply them to other baseline models, including LMNet \cite{LMnet} and Motionseg3D \cite{MotionSeg3D}. The results present in Tab. \ref{comparation} highlight their versatility and efficacy in improving the performance of both algorithms. Remarkably, our SIEM module significantly improves the performance of MotionSeg3D, achieving a +3.3\% increase in IoU.

As illustrated in Sec. \ref{sec:dataaug}, we evaluate different distributions and various ranges of $\Delta{t}$ in our distribution-based data augmentation. We select $\Delta{t}$ based on the designed distribution probability in each training iteration and used $\Delta{t}=1$ and $\Delta{t}=\text{max}$ during testing. The results are presented in Tab. \ref{aug}. Initially, we test different range values with an average distribution of $\Delta{t}$. When using a wider range of $\Delta{t}$, the performance deteriorates in the testing with $\Delta{t}=1$, while the performance was consistently better in the testing with $\Delta{t}=\text{max}$, indicating non-robustness to different inputs. We further examine the effect of different distribution probabilities for $\Delta{t}$ within the range $\Delta{t} \in [1, 2, 3]$, as it exhibits the highest robustness. Gradually increasing the proportion of $\Delta{t}=1$ during training, we achieve the best performance with distribution probabilities [0.5, 0.25, 0.25] for $\Delta{t}=[1, 2, 3]$, respectively.

\begin{table}[t]
\centering
\caption{Model inference time (ms) results.}
\resizebox{\linewidth}{!}
{\begin{tabular}{ccccc}
\toprule
InsMOS & MotionSeg-v1 & MF-MOS-v1 & MotionSeg-v2 & MF-MOS-v2 \\
\midrule
193.68 & 45.93 & \textbf{37.48} & 117.01 & 96.19 \\
\bottomrule
\end{tabular}}
\label{runtime}
\vspace{-.5cm}
\end{table}

%we conducted statistical analysis on the IoU metrics of the validation set with different ranges of time interval t, different probability distributions of t during training, and fixed t values during validation. The results from i, ii, iii, v, and vi indicate that using a larger time interval t during the inference stage yields better results compared to the original adjacent interval residual map sequences. This verifies that, under certain conditions, increasing the time interval of the residual maps can improve the model's segmentation capability for slowly moving objects. Compared to all experiments, it can be inferred that the probability distribution of t values used in iii yields the best results during training.

\subsection{Qualitative Analysis}
% \begin{figure*}[htbp]
%     \centering
%     \includegraphics[width=0.8\linewidth]{figs/fig2.png}
%    \caption{Dual-Branch, Dual-Head Architecture.  We enhance ME-MOS with two branches: an image branch capturing appearance and a residual map branch encoding temporal motion.  These branches merge using a multi-scale motion-guided attention module.  An image decoder with skip connections extracts features, followed by 2D-to-3D projection and further segmentation refinement via a point decoder. We adapt the SalsaNext network by replacing its pooling kernel with the Strip Average Pooling Layer (SAPL), to address dimension misalignment between range and residual maps.}
% \label{Framework}
% \end{figure*}
In order to more intuitively compare our algorithm with other SoTA algorithms, we perform a visual qualitative analysis on the SemanticKITTI dataset. As shown in Fig. \ref{compare}, both LMNet and MotionSeg3D have misjudgments of movable objects and missed determinations of moving objects. Compared with the SoTA algorithms, we can effectively remove the influence of movable objects through a model based on residual maps, and accurately capture moving objects.

\subsection{Runtime}
All the comparative experiments are performed on a single V100 GPU for inference. As shown in Tab. \ref{runtime}, in the one-stage comparison, MF-MOS-v1 outperforms other models. In the two-stage comparison, MF-MOS-v2 achieves real-time processing speed and surpasses Motionseg3D-v2 in terms of performance. 

\section{CONCLUSIONS}
This paper presents a dual-branch motion-based LiDAR moving object segmentation framework, a spatial-guided information enhancement module, and a distribution-based data augmentation method. Extensive experimental results demonstrate that 1) the framework MF-MOS in this study achieves the highest accuracy on both the validation and test sets, respectively, and 2) the proposed model demonstrates superior performance and generalization capabilities, making it applicable to other range-based methods.
\vspace{.5cm}

%%%%%%%%%%%%%%%%%%%%%%%%%%%%%%%%%%%%%%%%%%%%%%%%%%%%%%%%%%%%%%%%%%%%%%%%%%%%%%%%

%%%%%%%%%%%%%%%%%%%%%%%%%%%%%%%%%%%%%%%%%%%%%%%%%%%%%%%%%%%%%%%%%%%%%%%%%%%%%%%%

%%%%%%%%%%%%%%%%%%%%%%%%%%%%%%%%%%%%%%%%%%%%%%%%%%%%%%%%%%%%%%%%%%%%%%%%%%%%%%%%

\bibliographystyle{ieeetr}
\bibliography{bibtex/bib/IEEEexample}

%\begin{thebibliography}{99}

%\end{thebibliography}

\end{document}

% \maketitle
\thispagestyle{empty}
\pagestyle{empty}

%%%%%%%%%%%%%%%%%%%%%%%%%%%%%%%%%%%%%%%%%%%%%%%%%%%%%%%%%%%%%%%%%%%%%%%%%%%%%%%%

\section*{APPENDIX}

\subsection{Experiment Setups}

\subsubsection{Datasets}
SemanticKITTI-MOS \cite{LMnet} is a classical dataset for moving object segmentation tasks with standard labels about moving objects. Similarly to other approaches, we partitioned the dataset into three subsets. Furthermore, we augmented the dataset with semantic labels for movable objects and provided supervision for the main branch. To further provide meaningful data, we added the KITTI-road dataset for training phrases in some experiments.

In addition to conducting extensive experiments using the aforementioned dataset, we also performed validation experiments on Apollo\cite{apollo}. Similar to the sequences used in \cite{4d}, a subset of experiments conducted quantitative analysis on the Apollo dataset.

\subsubsection{Training Details and Evaluation Metric}
Our model was trained using the PyTorch deep learning framework on 2 NVIDIA Tesla A100 GPUs. The total number of training epochs was set to 150, with an initial learning rate of 0.01. The learning rate was decayed by a factor of 0.99. During training, a batch size of 8 was used on each GPU, and parameter optimization was performed using stochastic gradient descent (SGD) with a momentum of 0.9. During the training process, the total loss function includes the semantic-branch and motion-branch. Both semantic and motion branches used the weighted cross-entropy and Lov'asz-Softmax losses.
To valid our experiment result, we use intersection-over-union (IoU) \cite{IoU} of the moving objects as the evaluation metric.

\bibliographystyle{ieeetr}
\bibliography{bibtex/bib/IEEEexample}